\documentclass[times,10pt,twocolumn]{article}
\usepackage{latex8}
\usepackage{times}
\usepackage{balance}

\AtBeginDocument{%
  \providecommand\BibTeX{{%
    \normalfont B\kern-0.5em{\scshape i\kern-0.25em b}\kern-0.8em\TeX}}}

\usepackage{amssymb}
\usepackage{xspace}
\usepackage{caption}
\usepackage[skip=0cm,list=true,labelfont=bf]{subcaption}

\usepackage{soul}
\usepackage{lipsum,graphicx,multicol}
\usepackage{tabularx}

\usepackage{color, colortbl}
\definecolor{Gray}{gray}{0.9}
\usepackage[first=0,last=9]{lcg}

\usepackage[]{graphicx} 

\usepackage{algorithm}
\usepackage[noend]{algpseudocode}
\usepackage{nicefrac}
\providecommand{\keywords}[1]{\textbf{\textit{Index terms---}} #1}

\usepackage{amsmath}
\usepackage{algorithm}
\usepackage[noend]{algpseudocode}

\makeatletter
\def\BState{\State\hskip-\ALG@thistlm}
\renewcommand{\paragraph}{\@startsection{paragraph}{4}{\z@}%
  {0.5ex \@plus 2ex \@minus -3ex}
  {-1em}
  {\normalfont\normalsize\bfseries}}    

\makeatother

\begin{document}

\title{Transformer-Based Person Identification \\ via Wi-Fi CSI Amplitude and Phase Perturbations}
%
\author{Danilo Avola\textsuperscript{1}, Andrea Bernardini\textsuperscript{2}, Francesco Danese\textsuperscript{1}, Mario Lezoche\textsuperscript{3},\\
Maurizio Mancini\textsuperscript{1}, Daniele Pannone\textsuperscript{1}, and Amedeo Ranaldi\textsuperscript{1}\\
\textsuperscript{1}Department of Computer Science, Sapienza University of Rome\\
Via Salaria 113, 00198, Rome (RM), Italy\\
\{avola,danese,m.mancini,pannone,ranaldi\}@di.uniroma1.it\\
\textsuperscript{2}Fondazione Ugo Bordoni\\
Viale del Policlinico 147, 00161, Rome (RM), Italy\\
abernardini@fub.it\\
\textsuperscript{3}University of Lorraine, CNRS, CRAN\\
Nancy, F-54000, France\\
mario.lezoche@univ-lorraine.fr
}

\maketitle
\thispagestyle{empty}
%
\begin{abstract}
Wi-Fi sensing is gaining momentum as a non-intrusive and privacy-preserving alternative to vision-based systems for human identification. However, person identification through wireless signals, particularly without user motion, remains largely unexplored. Most prior wireless-based approaches rely on movement patterns, such as walking gait, to extract biometric cues. In contrast, we propose a transformer-based method that identifies individuals from Channel State Information (CSI) recorded while the subject remains stationary. CSI captures fine-grained amplitude and phase distortions induced by the unique interaction between the human body and the radio signal. To support evaluation, we introduce a dataset acquired with ESP32 devices in a controlled indoor environment, featuring six participants observed across multiple orientations. A tailored preprocessing pipeline, including outlier removal, smoothing, and phase calibration, enhances signal quality. Our dual-branch transformer architecture processes amplitude and phase modalities separately and achieves 99.82\% classification accuracy, outperforming convolutional and multilayer perceptron baselines. These results demonstrate the discriminative potential of CSI perturbations, highlighting their capacity to encode biometric traits in a consistent manner. They further confirm the viability of passive, device-free person identification using low-cost commodity Wi-Fi hardware in real-world settings.
\end{abstract}


\section{Introduction}
Person identification plays a crucial role in a wide range of applications, including access control, surveillance, personalized services, human-computer interaction within smart environments, and many others. Biometric identification systems can be broadly categorized into two classes: \textbf{proximal biometrics}, which include both contact-based and near-contact modalities such as fingerprint recognition, facial authentication, and iris scanning, and \textbf{remote sensing approaches}, which aim to recognize individuals from a distance, without requiring physical contact or explicit cooperation, such as gait recognition. Proximal systems often rely on specialized sensors and controlled conditions. For instance, fingerprint and palmprint scanners require direct contact, while face and iris recognition systems depend on precise user alignment and favorable illumination. Although these technologies are well-established and accurate \cite{10.1145/954339.954342,BOWYER2008281,7380482}, these constraints can raise concerns related to user compliance, susceptibility to spoofing, and limitations in scalability, especially in high-traffic or privacy-sensitive environments. In contrast, remote sensing methods overcome many of these limitations by enabling passive or semi-passive identification. Among them, gait recognition has become a popular contactless biometric that leverages spatiotemporal patterns in human locomotion \cite{8528404,AVOLA2024103643}. Other techniques include thermal imaging and millimeter-wave radar, which can capture distinctive physiological or behavioral traits without requiring physical contact, user cooperation, or constrained acquisition conditions.

Moving within the domain of radio frequency (RF) signals, it is important to note that Wi-Fi sensing has evolved over the past two decades into a mature and versatile discipline, extending far beyond its original role in wireless communication. Today, Wi-Fi-based systems are increasingly employed to perform tasks that were traditionally the exclusive domain of computer vision, including human localization, fall detection, gesture recognition, occupancy estimation, activity classification, and many others. Unlike vision-based systems, Wi-Fi sensing offers key advantages such as resilience to occlusions, variable lighting conditions, and camera placement constraints. It also operates without requiring additional infrastructure beyond standard wireless access points, enabling fully device-free and privacy-preserving sensing. These characteristics make Wi-Fi an attractive alternative for ambient intelligence scenarios. Among the many applications of Wi-Fi sensing, person identification remains one of the most underexplored. While a few recent studies have explored the possibility of distinguishing individuals based on dynamic signal variations, primarily through gait patterns or gesture-induced perturbations in Wi-Fi signals \cite{7536315}, the task of recognizing a person in a fully static and passive scenario remains largely unaddressed in the literature. Recent findings suggest that the propagation of Wi-Fi signals through the human body is modulated by subject-specific internal factors, including the anatomical arrangement of organs, tissue composition, and water content \cite{doi:10.1142/S0129065722500150,9730862,doi:10.1177/10692509251339913}. These physiological characteristics subtly distort the channel response in ways that can serve as biometric signatures, even in the absence of motion. This opens up the possibility for contactless, device-free person identification based solely on how the human body inherently interacts with the surrounding electromagnetic field. A key enabler of these capabilities is Channel State Information (CSI), which provides fine-grained amplitude and phase data across multiple subcarriers of the Wi-Fi signal. CSI has been successfully employed in a variety of sensing tasks, including gesture recognition, activity detection, and occupancy estimation \cite{9516988,10.1145/3310194}. However, its application to static person identification remains extremely limited. Most prior work in this space depends on motion-induced signal dynamics, such as those generated during walking or gesturing, making identification in the absence of movement a significantly underexplored and challenging problem.

To address this gap, we propose a novel transformer-based approach for static person identification using Wi-Fi CSI. Our method extracts identity-discriminative features from subtle amplitude and phase perturbations, even when the subject remains completely stationary. Given the absence of publicly available datasets for this setting, we collected a custom dataset using low-cost ESP32 devices in a controlled indoor environment, capturing CSI data from six individuals across multiple orientations. Although limited in scale, this dataset establishes an initial benchmark for evaluating passive Wi-Fi-based biometric identification. We further introduce a lightweight yet effective dual-branch transformer architecture, which processes amplitude and phase streams independently before merging them via late fusion. The model is designed to capture cross-temporal dependencies without relying on convolutional or recurrent mechanisms. Experimental results on our dataset show that it achieves a classification accuracy of 99.82\%, outperforming standard baselines such as convolutional neural networks (CNNs) and multilayer perceptrons (MLPs). In summary, this work makes the following contributions:
\begin{itemize}
  \item We position Wi-Fi CSI-based identification within the broader landscape of biometric technologies, highlighting the gap in static, passive settings.
  \item We introduce a new dataset for static person identification, acquired with low-cost, commodity Wi-Fi hardware under a controlled acquisition protocol.
  \item We propose a novel dual-branch transformer architecture that independently models amplitude and phase dynamics through separate attention pathways, demonstrating superior performance compared to standard CNN and MLP baselines in a challenging, static, device-free biometric identification setting.
\end{itemize}

\section{Related work} \label{sec:related_work}
\begin{figure*}[t]
\centering
\includegraphics[width=0.9\linewidth]{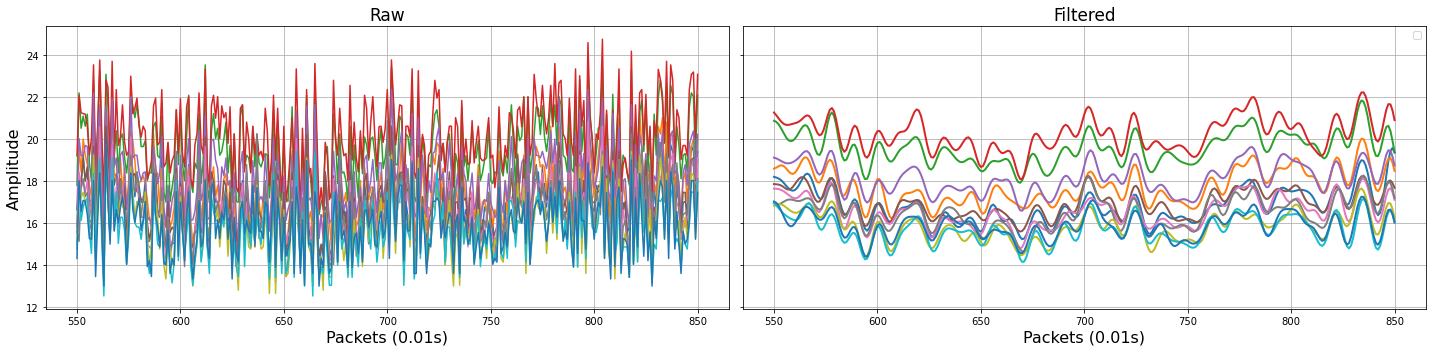}
\caption{Illustrative example of amplitude signals from 11 representative Wi-Fi subcarriers over a 3-second acquisition window, shown before and after the denoising process applied during the preprocessing pipeline to enhance signal quality.
}
\label{fig:amp-prep}
\end{figure*}
Wi-Fi sensing has emerged as a versatile paradigm for extracting human-related information from ambient radio signals. While this field spans a wide range of applications and possible categorizations, for the purposes of this work, we distinguish between two broad lines of research: Wi-Fi sensing for general human-centered tasks, such as gesture recognition or activity monitoring, and Wi-Fi-based systems specifically designed for biometric person identification. This separation allows us to more clearly contextualize our contribution with respect to both the sensing methodologies and the biometric objectives of the current literature.

As regards the first category, over the past decade, Wi-Fi sensing has been widely adopted in a broad range of human-centered applications, including gesture recognition, activity monitoring, indoor localization, fall detection, occupancy estimation, and many others. Early systems primarily relied on the Received Signal Strength Indicator (RSSI) to infer user presence or movement. However, RSSI is characterized by low spatial resolution and high sensitivity to multipath effects, making it unreliable in cluttered or dynamic environments. These limitations motivated the transition to CSI, which provides more detailed and robust measurements by capturing amplitude and phase variations across multiple subcarriers \cite{CSI_RSSI,RSSI_indoorloc,RSSI_reliability_indoorloc}. Recent advances in gesture and activity recognition have demonstrated the potential of CSI-based systems to capture fine-grained motion patterns over time. These approaches typically treat the temporal evolution of amplitude and phase as a multivariate time series, from which meaningful features can be extracted to characterize human behavior. One notable example is Widar3.0 \cite{9516988}, which performs cross-domain gesture recognition by learning velocity-invariant representations that generalize across locations and environments. Other methods leverage deep learning architectures such as Convolutional Neural Networks (CNNs), Long Short-Term Memory networks (LSTMs), and, more recently, Transformer-based models to capture spatiotemporal dependencies within CSI sequences \cite{10.1145/3310194,9516988,phase_noise}. Despite their success in modeling dynamic behaviors, these techniques inherently depend on user motion and are therefore unsuitable for passive identification scenarios, where the subject remains stationary.

As regards the second category, the task of identifying individuals through Wi-Fi CSI remains substantially underexplored. Most existing approaches are dynamic and gait-based, such as WiWho~\cite{WiWho_gait_decisionTree} and Hampel+PCA~\cite{Gait_hampel_pca}, which require subjects to walk along predefined trajectories. Although effective, these methods are unsuitable for passive or stationary identification, and their performance is often environment-dependent. A few static approaches have attempted to identify individuals based on CSI signal distortions induced by the human body at rest. WiPIN~\cite{WiPIN} is one of the earliest systems in this category. It uses handcrafted statistical features and classical classifiers to distinguish users, but lacks robustness and scalability. Moreover, the absence of public datasets and standardized acquisition protocols across the literature makes fair comparison difficult and hinders reproducibility \cite{EdgeWiFiSensing2022}. Unlike prior works, which primarily rely on user motion or gait-induced signal variations, our method addresses the more challenging task of person identification in a fully static and passive setting, where the subject remains completely still and performs no explicit action. This scenario is particularly relevant for ambient intelligence applications that demand unobtrusive and privacy-preserving solutions, such as smart homes, assisted living environments, or secure access systems in sensitive contexts. To support systematic experimentation, we design a reproducible acquisition setup based on low-cost ESP32 devices \cite{ESP32_tool}, configured to operate in a controlled indoor environment. Our data collection protocol involves multiple recording sessions across different body orientations, ensuring consistency while capturing sufficient intra-subject variability. Given the absence of publicly available benchmarks in this setting, our dataset also serves as an important resource for future research on static RF-based biometrics. We further introduce a transformer-based dual-branch architecture that processes CSI amplitude and phase components independently through parallel attention-based encoding streams. This design allows the model to exploit complementary information from both modalities while maintaining architectural simplicity and computational efficiency. The two feature representations are later fused for classification via a lightweight decision head, enabling the model to capture fine-grained biometric traits embedded in the wireless channel. To the best of our knowledge, this is the first work to demonstrate high-accuracy person identification from CSI under strictly static, fully device-free conditions using deep learning.

\section{Proposed Method} \label{sec:proposed-method}
In wireless communications, CSI characterizes how a transmitted signal is modified by the physical wireless channel due to phenomena such as multipath propagation, attenuation, and phase shifting caused by environmental and anatomical factors. The channel can be modeled as:
\begin{equation}
Y = H X + \eta,
\end{equation}
where \( Y \in \mathbb{C} \) is the received signal, \( X \in \mathbb{C} \) is the known transmitted signal (often a pilot symbol), \( H \in \mathbb{C} \) represents the channel frequency response (i.e., the CSI), and \( \eta \in \mathbb{C} \) denotes additive noise. This model is applied per subcarrier in Orthogonal Frequency-Division Multiplexing (OFDM) systems. Since \( X \) is known and \( Y \) is measured, the channel estimate for a subcarrier \( k \) can be computed as:
\begin{equation}
\hat{H}_k = \frac{Y_k}{X_k}.
\end{equation}
\noindent This complex number encodes both amplitude attenuation and phase shift introduced by the channel. The corresponding amplitude \( A_k \) and phase \( \phi_k \) can be extracted from the real and imaginary parts of \( \hat{H}_k \) as follows:
\begin{equation}
A_k = \sqrt{\Re(\hat{H}_k)^2 + \Im(\hat{H}_k)^2},
\end{equation}
\begin{equation}
\phi_k = \text{atan2}\big(\Im(\hat{H}_k), \Re(\hat{H}_k)\big).
\end{equation}
\noindent These amplitude and phase components serve as the input to all subsequent CSI-based sensing or inference stages.

\subsection{Signal Preprocessing}
To enhance the quality and reliability of the CSI data for Wi-Fi sensing applications, we implemented a comprehensive signal preprocessing pipeline. The raw amplitude and phase data are structured as matrices of size \( T \times K \), where \( T \) denotes the number of time samples and \( K \) the number of subcarriers.
To reduce high-frequency noise and data dimensionality, we applied a temporal mean reduction technique. This involved averaging every two consecutive time samples, effectively halving the temporal resolution and resulting in a matrix of size \( \frac{T}{2} \times K \). Mathematically, for each subcarrier \( k \) and time index \( t \in [0, 1, \dots, \frac{T}{2} - 1] \), the reduced signal \( X^{(r)}_{t,k} \) is computed as:

\begin{equation}
X^{(r)}_{t,k} = \frac{1}{2} \left( X_{2t,k} + X_{2t+1,k} \right).
\end{equation}

\noindent This step serves to smooth out rapid fluctuations and reduce computational complexity in subsequent analyses. The amplitude component of CSI data is prone to outliers and sudden fluctuations due to hardware instability, environmental interference, and human motion. To mitigate such distortions, we employed the Hampel filter \cite{hampel_ID, Gait_hampel_pca}, a robust statistical technique that identifies outliers based on the Median Absolute Deviation (MAD) within a sliding window. Let \( X_w \) denote a segment of the amplitude time series defined by a sliding window of fixed length \( w \), where \( w \) is an odd number. The MAD within this window is computed as:

{\small
\begin{equation}
MAD_w = \text{Median} \left\{\, \left|\, x - \text{median}(X_w)\, \right| \, \, \, \forall x \in X_w \right\}.
\end{equation}
}
\noindent \hspace{-0.702em} The central point in the window, \( x_{w/2} \), is considered an outlier if it deviates significantly from the median of the window by more than \( \beta \) times the MAD threshold:

\begin{equation}
\left| x_{w/2} - \text{median}(X_w) \right| > \beta \cdot MAD_w.
\end{equation}
\begin{figure}[t]
\centering
\includegraphics[width=0.95\linewidth]{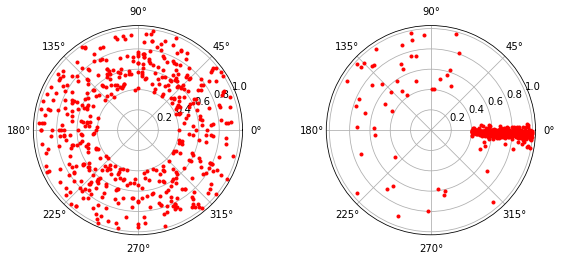}
\caption{Phase values of a single subcarrier before (left) and after calibration (right), illustrating the correction of random offsets and noise.}
\label{fig:calibration}
\end{figure}

\noindent Here, \( \beta \) is an application-specific threshold. This outlier detection procedure is applied independently to the amplitude time series of each subcarrier. To preserve temporal consistency, any detected outlier is replaced using exponential smoothing over previous values in the window. The smoothed replacement is computed as:
{\small
\begin{equation}
s_0 = x_0, \\ 
\quad \, s_{i} = \alpha x_i + (1 - \alpha) s_{i-1} \quad \, \forall i \in [1, w/2],
\end{equation}
}
\hspace{-0.7em} where \( x_i \) is the amplitude observation, \( s_i \) is the smoothed estimate, and \( \alpha \in (0, 1) \) is a smoothing factor controlling the influence of past values. This robust strategy ensures that isolated outliers are replaced while maintaining the temporal structure of the signal. Following outlier correction and smoothing, the amplitude time series are further denoised using a low-pass Butterworth filter \cite{Butterworth1930}. This filtering step is motivated by the need to suppress residual high-frequency noise while preserving the essential slow-varying signal characteristics caused by human motion and environmental dynamics. This operation is applied independently to the amplitude series of each subcarrier, producing a smoother amplitude profile that is well-suited for the downstream classification task, as illustrated in Fig.~\ref{fig:amp-prep}.

The phase component of CSI data is often distorted by hardware-induced errors \cite{phase_noise} such as Carrier Frequency Offset (CFO) and Sampling Frequency Offset (SFO), leading to linear and constant phase shifts across subcarriers. To correct these distortions, it is common to model the actual phase \( \hat{\phi}_i \) of subcarrier \( k_i \) as:

{\small
\begin{equation}
s_0 = x_0,
\hat{\phi}_i = \phi_i - 2\pi \frac{k_i}{N} \delta + \beta + \eta,
\end{equation}
}
\hspace{-0.6em} where \( \phi_i \) is the true phase, \( \delta \) represents the timing offset due to SFO, \( \beta \) is the constant phase offset from CFO, \( \eta \) denotes measurement noise, and \( N \) is the total number of subcarriers. To estimate and remove the linear trend introduced by SFO and CFO, we applied a linear regression procedure that models the raw phase response of each packet as a function of the subcarrier index \cite{Phase_cal1, Phase_cal3}. This linear component arises due to hardware imperfections in the transmitter and receiver, and manifests as a frequency-dependent phase distortion that must be corrected before meaningful analysis. Specifically, we estimated the slope \( a \) and intercept \( b \) of the regression line using only the phase values at the endpoints of the subcarrier range. This approach provides a computationally efficient and sufficiently accurate approximation of the linear trend affecting the phase, which can then be subtracted to obtain a drift-compensated phase profile:
\begin{equation}
a = \frac{\hat{\phi}_n - \hat{\phi}_1}{k_n - k_1}, \quad b = \frac{1}{n} \sum_{j=1}^{n} \hat{\phi}_j,
\end{equation}

\noindent here, \( \hat{\phi}_1 \) and \( \hat{\phi}_n \) are the sensed phases of the first and last subcarriers, respectively, and \( k_1 \) and \( k_n \) are their corresponding indices. This simplified approach assumes a linear phase distortion across subcarriers, which is a reasonable approximation given the nature of CFO and SFO effects. The calibrated phase \( \Phi_i \) for each subcarrier was then obtained by subtracting the estimated linear trend:
\begin{equation}
\Phi_i = \hat{\phi}_i - a k_i - b.
\end{equation}

\noindent Fig.~\ref{fig:calibration} demonstrates how the phases of one subcarrier are sanitized with the procedure. This calibration removes the deterministic linear phase errors, enhancing the reliability of phase information for subsequent sensing tasks.

\subsection{Transformer-Based Model}
\begin{figure}
    \centering    
    \includegraphics[width=0.9\linewidth]{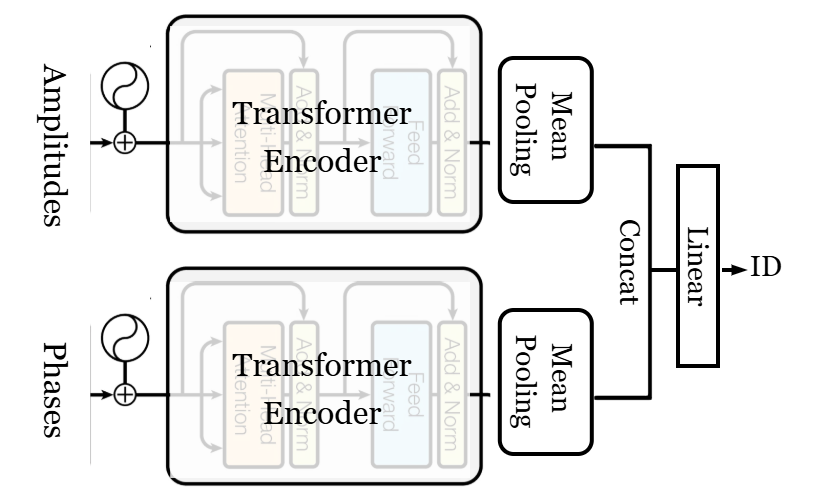}
    \caption{Architecture of the proposed Transformer-based model for person identification.}
    \label{fig:model}
\end{figure}
\begin{figure*}[t]
    \centering    \includegraphics[width=0.9\linewidth]{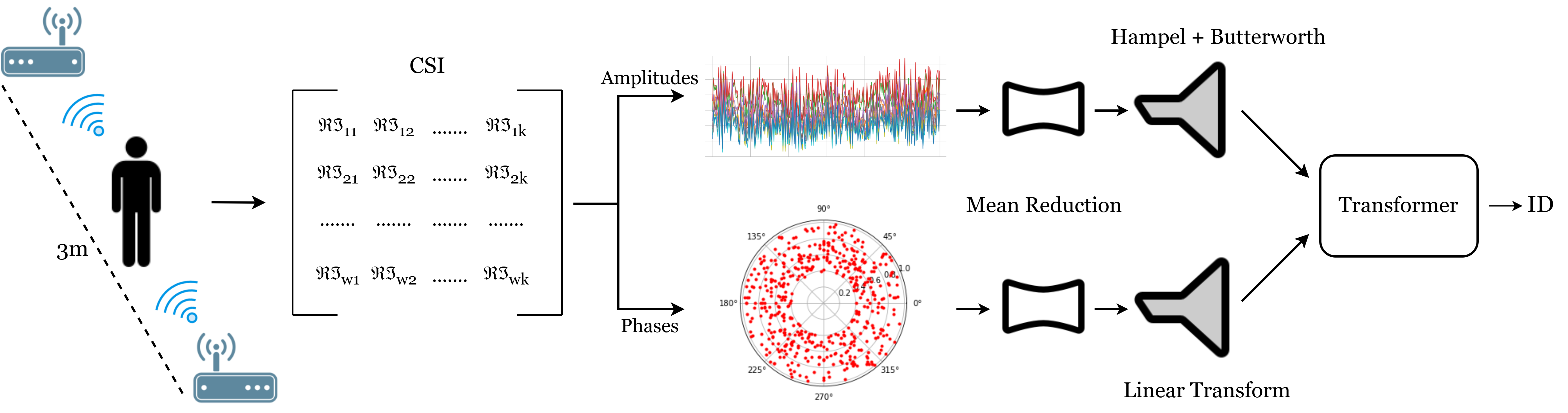}
    \caption{Overview of the complete processing pipeline of the proposed Wi-Fi-based person identification system, illustrating each stage from signal acquisition to final classification.}
    \label{fig:pipeline}
\end{figure*}

Inspired by the original Transformer architecture \cite{vaswani2023attentionneed}, we propose a streamlined attention-based model for learning representations from the preprocessed amplitude and phase information, as illustrated in Fig.~\ref{fig:model}. The architecture handles these modalities separately using two parallel encoders. Each input is represented as a sequence of $W = T/2$ time steps and $K$ subcarriers, forming two matrices: $\mathbf{X}_{\text{amp}} \in \mathbb{R}^{W \times K}$ for amplitude and $\mathbf{X}_{\text{ph}} \in \mathbb{R}^{W \times K}$ for phase. Each of these is projected into a $d_{\text{model}}$-dimensional embedding space through a learnable linear transformation:
\begin{equation}
\mathbf{H}_{\text{amp}} = \mathbf{X}_{\text{amp}} \mathbf{W}_{\text{in,amp}} + \mathbf{b}_{\text{in,amp}},
\end{equation}
\begin{equation}
\mathbf{H}_{\text{ph}} = \mathbf{X}_{\text{ph}} \mathbf{W}_{\text{in,ph}} + \mathbf{b}_{\text{in,ph}},
\end{equation}
where $\mathbf{W}_{\text{in,amp}}, \mathbf{W}_{\text{in,ph}} \in \mathbb{R}^{K \times d_{\text{model}}}$. To encode the temporal order, sinusoidal positional encoding is added to $\mathbf{H}_{\text{amp}}$ and $\mathbf{H}_{\text{ph}}$. These positionally encoded sequences are then passed through a single transformer encoder layer. The transformer layer consists of a multi-head self-attention mechanism followed by a feed-forward network, each with residual connections and layer normalization.
Let $\mathbf{H}$ denote either $\mathbf{H}_{\text{amp}}$ or $\mathbf{H}_{\text{ph}}$. The self-attention output is computed as:
\begin{equation}
\text{Attention}(\mathbf{Q}, \mathbf{K}, \mathbf{V}) = \text{softmax}\left(\frac{\mathbf{Q} \mathbf{K}^\top}{\sqrt{d_{\text{model}}}}\right) \mathbf{V},
\end{equation}
\begin{equation}
\mathbf{Q} = \mathbf{H} \mathbf{W}^Q, \ \mathbf{K} = \mathbf{H} \mathbf{W}^K, \ \mathbf{V} = \mathbf{H} \mathbf{W}^V,
\end{equation}
where $\mathbf{W}^Q, \mathbf{W}^K, \mathbf{W}^V \in \mathbb{R}^{d_{\text{model}} \times d_{\text{model}}}$ are learned projection matrices. The multi-head version concatenates multiple such heads and projects them back into $d_{\text{model}}$ dimensions. The multi-head self-attention output is then combined with the input through a residual connection and normalized:
\begin{equation}
\mathbf{H}^{(1)} = \text{LayerNorm}(\mathbf{H} + \text{Dropout}(\text{MultiHead}(\mathbf{H}))).
\end{equation}

The attention output is subsequently passed through a position-wise feed-forward layer with ReLU activation:
\begin{equation}
\text{FFN}(\mathbf{H}^{(1)}) = \max(0, \mathbf{H}^{(1)} \mathbf{W}_1 + \mathbf{b}_1) \mathbf{W}_2 + \mathbf{b}_2,
\end{equation}
where $\mathbf{W}_1 \in \mathbb{R}^{d_{\text{model}} \times d_{\text{ff}}}$ and $\mathbf{W}_2 \in \mathbb{R}^{d_{\text{ff}} \times d_{\text{model}}}$. The feed-forward sublayer also employs a residual connection followed by layer normalization:
\begin{equation}
\mathbf{Z} = \text{LayerNorm}(\mathbf{H}^{(1)} + \text{Dropout}(\text{FFN}(\mathbf{H}^{(1)}))),
\end{equation}
after the transformer encoder layer, the output sequences $\mathbf{Z}_{\text{amp}} \in \mathbb{R}^{W \times d_{\text{model}}}$ and $\mathbf{Z}_{\text{phase}} \in \mathbb{R}^{W \times d_{\text{model}}}$ are aggregated via mean pooling across the temporal dimension:
\begin{equation}
\mathbf{z}_{\text{amp}} = \frac{1}{W} \sum_{i=1}^{W} \mathbf{Z}_{\text{amp}}[i],
\end{equation}
\begin{equation}
\mathbf{z}_{\text{ph}} = \frac{1}{W} \sum_{i=1}^{W} \mathbf{Z}_{\text{ph}}[i].
\end{equation}

\noindent The resulting vectors are concatenated:
\begin{equation}
\mathbf{z} = [\mathbf{z}_{\text{amp}}; \mathbf{z}_{\text{ph}}] \in \mathbb{R}^{2 d_{\text{model}}},
\end{equation}
and then passed through a linear classification head to produce the final output logits over $N$ classes:
\begin{equation}
\hat{\mathbf{y}} = \mathbf{z} \mathbf{W}_{\text{out}} + \mathbf{b}_{\text{out}}, \quad \mathbf{W}_{\text{out}} \in \mathbb{R}^{2 d_{\text{model}} \times N}.
\end{equation}

This lightweight architecture fully leverages the self-attention mechanism to model dependencies across time and subcarriers, while avoiding the use of convolutional or recurrent layers. The residual connections enable gradient flow during training, while layer normalization stabilizes the overall learning process. The mean pooling operation enables global aggregation of the temporal sequence without introducing any additional learnable parameters, ensuring simplicity and computational efficiency throughout.

\section{Experimental Results and Discussion}
Due to the absence of publicly available CSI datasets specifically designed for person identification, a new dataset was constructed. Unlike prior work relying on gait dynamics, our objective was to identify individuals while standing still. CSI data were collected using two ESP32 microcontrollers \cite{ESP32_tool} configured as transmitter (TX) and receiver (RX), spaced 3 meters apart and connected to laptops for continuous data recording. Six human subjects (four males, two females) participated in the acquisition process, each standing in a fixed indoor environment that remained unchanged throughout the sessions. All electronic devices, including the mobile phones of participants, were removed from the room to eliminate potential sources of electromagnetic interference. These precautions were taken to minimize any bias that could introduce spurious correlations between the environment and the data associated with a specific individual. This ensured that background noise remained consistent across all classes, thus allowing the model to learn biometric features of the person rather than detecting foreign artifacts. An overview of the entire Person-ID system is shown in Fig.~\ref{fig:pipeline}.

\begin{figure}[h]
  \centering
  \begin{minipage}[t]{0.5\linewidth}
    \centering
    \includegraphics[width=\linewidth]{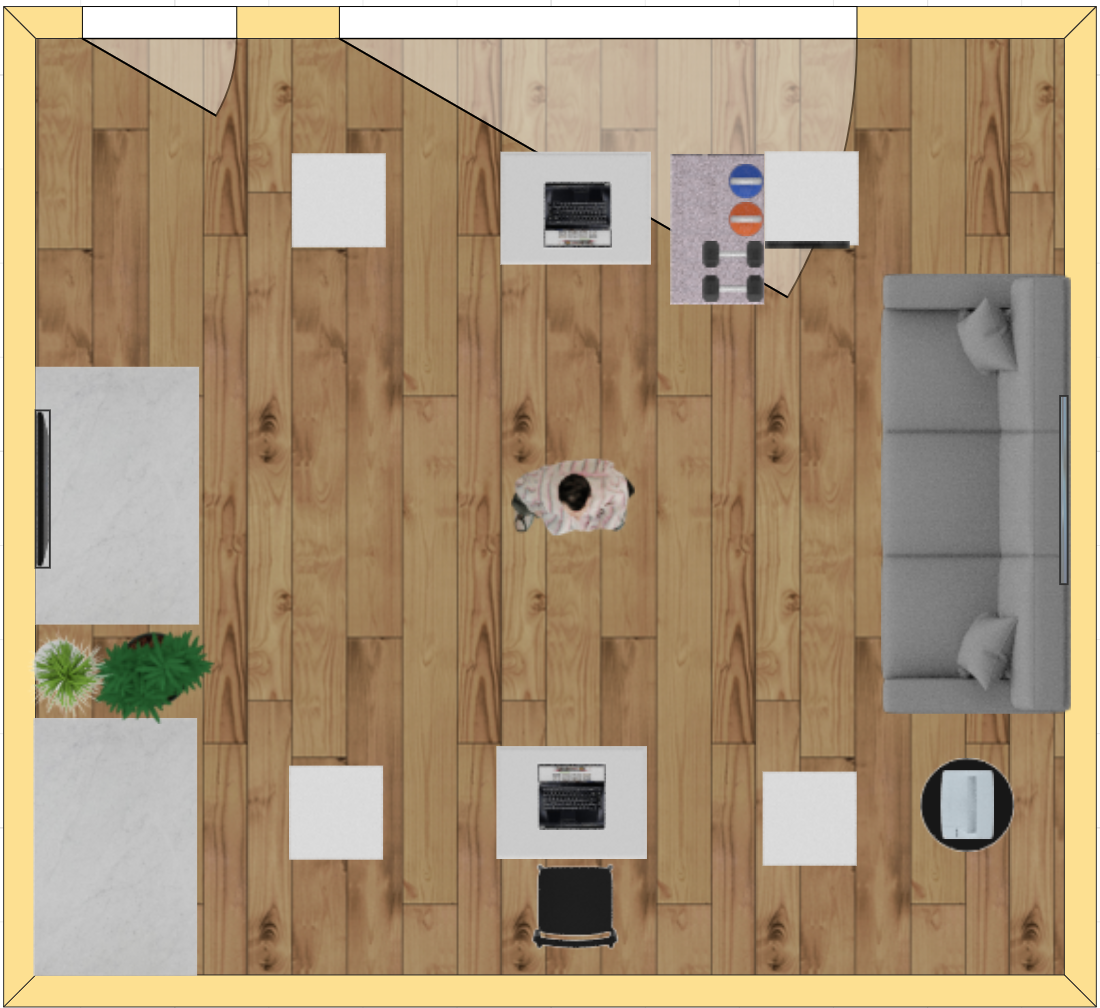}
  \end{minipage}
  \hfill
  \begin{minipage}[t]{0.45\linewidth}
    \centering    \includegraphics[width=0.9\linewidth]{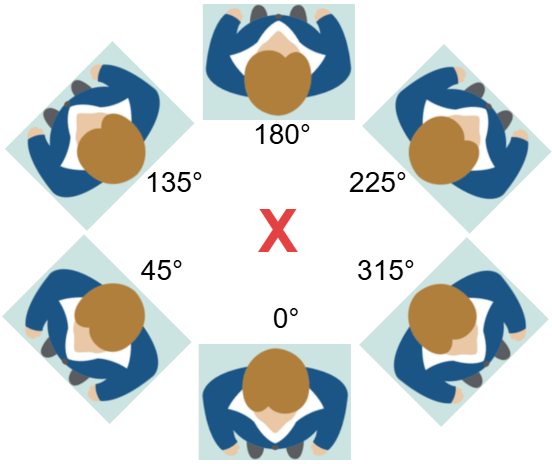}
  \end{minipage}
  \caption{A rendering of the data acquisition room (left) and an overview of the 6 different directions faced by the person at each round (right).}
  \label{fig:sidebyside}
\end{figure}

The demographics of the participants included two female subjects (average: 177 cm, 66 kg, 57 years) and four male subjects (average: 185 cm, 84 kg, 30 years). All measurements were taken under identical environmental conditions to ensure consistency. While the dataset is limited in scale, it was intentionally designed to serve as a solid and reproducible starting point for investigating static, device-free person identification from Wi-Fi signals. To the best of our knowledge, it represents a unique contribution in this setting, offering high-quality, multimodal CSI data acquired under controlled and well-documented conditions. Despite its minimal size, the dataset enables rigorous validation and lays the groundwork for future studies in this underexplored domain. Each subject was instructed to stand at the center point along the line-of-sight between TX and RX (i.e., 1.5 meters from both antennas) for six acquisition rounds, corresponding to six different orientations: $0^\circ$, $45^\circ$, $135^\circ$, $180^\circ$, $225^\circ$, and $315^\circ$. For each orientation, participants remained stationary and breathed normally for 2 minutes and 30 seconds. Slight movements such as speaking, coughing, scratching or posture shifts were tolerated. A view from above of the room structure, furniture position, and body orientations is shown in Fig. \ref{fig:sidebyside}.

CSI packets were sampled at 100 Hz, resulting in 90,000 samples per subject over the full duration (6 orientations $\times$ 150 seconds) recording a total of 52 subcarriers. Additionally, a CSI session of the empty room was recorded for comparison purposes. Each of the six time series was divided sequentially, with the first 70\% of samples used for training, the subsequent 10\% for validation, and the final 20\% for testing. After the split, to prepare the data for model input, we applied a sliding window of 100 packets (corresponding to 1 second) across each sequence. A 50\% overlap between consecutive windows was employed, which serves both as a form of data augmentation and as a means to reduce redundancy and computational overhead compared to a one-step sliding window.

\subsection{Training, Baseline, and Results}
Model deployment and training were conducted on PyTorch, using Adam optimizer with a learning rate of 0.001, cross-entropy loss, and a batch size of 32. Early stopping with a patience of 10 epochs was applied based on validation accuracy, with training performed for a maximum of 50 epochs. The Hampel identifier window size was set to 15, with $\beta = 3$ and $\alpha = 0.8$. The low-pass Butterworth filter was of $5^{th}$ order with cutoff frequency = 10 hz. For the transformer hyperparameters, we set the dropout rate to 0.2, $d_{\text{model}} = 32$ with 4 attention heads, $d_{\text{ff}} = 64$, and $T = 100$ corresponding to 100 packets. We compared our proposed approach against three baseline models, all preceded by the same preprocessing methods explained in Section~\ref{sec:proposed-method}. 

To contextualize the performance of our proposed method, we compare it against three baseline architectures representative of common approaches to signal classification. These baselines were selected to assess the effectiveness of different modeling strategies, ranging from non-temporal to convolutional and hybrid designs, when applied to CSI-based person identification. The first baseline is a simple MLP. In this configuration, amplitude and phase windows are first averaged along the temporal dimension, effectively discarding temporal dependencies. The resulting two feature vectors are concatenated and passed through an MLP consisting of a single hidden layer of size \( d_{\text{model}} \), equipped with ReLU activation, dropout regularization, and a final linear projection to the output classes. The second baseline is a CNN composed of three convolutional layers, each with 32 feature maps and interleaved with max-pooling operations. The amplitude and phase matrices are treated as separate input channels and fed directly into the network. After the final convolutional layer, the feature maps are flattened and passed through the same MLP head used in the previous baseline. The third baseline is a hybrid model that combines the transformer encoder with the CNN structure. In this variant, the output sequences from the two transformer encoder branches (one for amplitude and one for phase) are concatenated along the channel dimension and passed to the same CNN architecture described above. This design allows us to test whether local convolutional filtering can further enhance the sequence-level representations learned by the transformer modules, particularly in capturing spatial dependencies across feature dimensions.
\begin{figure}[t]
    \centering
    \includegraphics[width=0.95\linewidth]{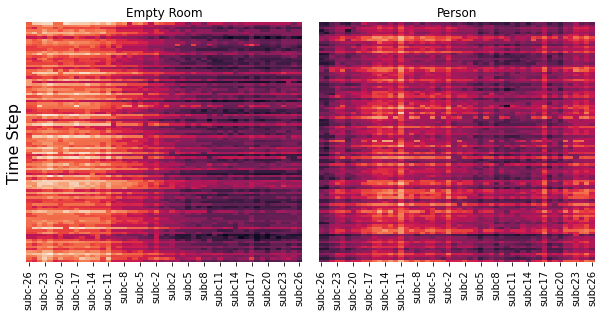}
    \caption{Heatmap of CSI amplitudes (each subcarrier over time) of an empty room, versus the same room with a subject standing in the middle of the scene.}
    \label{fig:emptyperson}
\end{figure}

Regarding the results, Fig.~\ref{fig:emptyperson} shows heatmaps of CSI amplitude values over a 2-second interval for all subcarriers, comparing two conditions: an empty room and a scenario where a person is standing between the transmitter and receiver. In the empty-room case, the amplitude heatmap displays a stable temporal structure, characterized by a smooth and consistent decrease in amplitude from lower to higher subcarrier indices. This attenuation pattern is likely due to hardware and path-loss effects and remains steady across time frames. Conversely, when a person is present in the environment, the amplitude distribution becomes significantly more irregular and dynamic. These perturbations, induced by the subject’s body interacting with the wireless channel, are highly distinctive and person-specific. It is precisely this type of fine-grained variation that our system is designed to capture and leverage for reliable static person identification based on CSI. For evaluation on the test set, the standard classification metrics of accuracy, precision, recall, and F1-score were computed, with the last three as macro-averages across classes. Given true positives (TP), false positives (FP), and false negatives (FN), the metrics are defined as:
\begin{equation}
\text{Precision} = \frac{TP}{TP + FP},
\end{equation}
\begin{equation}
\text{Recall} = \frac{TP}{TP + FN},
\end{equation}
\begin{equation}
\text{F1} = 2 \cdot \frac{\text{Precision} \cdot \text{Recall}}{\text{Precision} + \text{Recall}}.
\end{equation}

\noindent Empirically, we found that the suggested transformer-only model achieved superior performance and generalization compared to these alternatives, despite its simple architecture. Table~\ref{tab:performance} displays the evaluation results, where the proposed model reaches a global Accuracy of 99.82\% for six identities. Notably, even the temporal-agnostic MLP baseline yields strong results, with all metrics around 94.40\%, suggesting that the sanitized CSI data contain highly discriminative features relevant to person-ID. We also experimented with using a learnable classification token (\texttt{[CLS]}) as the sequence aggregator instead of mean pooling, but observed a slight performance drop, and thus retained the simpler averaging strategy in the final model.
\begin{table}[t]
\centering
\small
\begin{tabular}{l|c|c|c|c}
\hline
\textbf{Model} & \textbf{Accuracy} & \textbf{F1} & \textbf{Precision} & \textbf{Recall} \\
\hline
Transformer     & \textbf{99.82}\% & \textbf{99.81}\% & \textbf{99.82}\% & \textbf{99.81}\% \\
Trans + CNN     & 99.15\% & 99.16\% & 99.17\% & 99.15\% \\
CNN             & 98.72\% & 98.71\% & 98.72\% & 98.71\% \\
MLP             & 94.36\% & 94.44\% & 94.50\% & 94.40\% \\
\hline
\end{tabular}
\caption{Performance comparison of models on the test set. F1, precision, and recall are macro-averaged.}
\label{tab:performance}
\end{table}

\section{Conclusion}
This work presents a transformer-based approach for person identification using Wi-Fi CSI, leveraging subtle amplitude and phase distortions induced by human presence under static conditions. Unlike prior methods that depend on user motion or behavioral dynamics, our system operates in a fully passive manner, offering a non-intrusive and privacy-preserving alternative to vision-based techniques. To support reproducibility and consistent evaluation, we introduce a rigorous acquisition protocol based on low-cost ESP32 hardware, capturing CSI data from six individuals positioned in multiple static orientations. A robust preprocessing pipeline, including temporal averaging, Hampel filtering, Butterworth denoising, and phase calibration, ensures the extraction of clean and informative features from the raw signal traces. Our proposed transformer architecture processes amplitude and phase information through parallel encoding branches, enabling the model to capture complementary temporal patterns from each modality. The system achieves 99.82\% classification accuracy and comparable performance across other standard evaluation metrics, outperforming CNN and MLP baselines while maintaining a lightweight design with low computational complexity. These results demonstrate that, even in the absence of motion, CSI perturbations contain discriminative biometric signatures suitable for accurate and real-time person identification. Overall, this study confirms the feasibility and effectiveness of RF-based biometric sensing using readily available, low-cost Wi-Fi hardware. Future directions include extending the dataset to a larger and more diverse subject pool, investigating cross-environment generalization capabilities, and exploring setups involving multiple antennas transmitting from different spatial positions to further enrich signal diversity and robustness. In addition, integrating temporal stability analysis and testing on more realistic deployment conditions could help validate long-term applicability. We believe this work lays the foundation for developing practical, contactless identification systems based entirely on ambient radio signals.

\section*{Acknowledgments}
This work was supported by ``Smart unmannEd AeRial vehiCles for Human likE monitoRing (SEARCHER)'' project of the Italian Ministry of Defence within the PNRM 2020 Program (Grant Number: PNRM a2020.231); ``EYE-FI.AI: going bEYond computEr vision paradigm using wi-FI signals in AI systems'' project of the Italian Ministry of Universities and Research (MUR) within the PRIN 2022 Program (Grant Number: 2022AL45R2) (CUP: B53D23012950001); MICS (Made in Italy – Circular and Sustainable) Extended Partnership and received funding from Next-Generation EU (Italian PNRR – M4 C2, Invest 1.3 – D.D. 1551.11-10-2022, PE00000004) (CUP MICS B53C22004130001); and “Enhancing Robotics with Human Attention Mechanism via Brain-Computer Interfaces” Sapienza University Research Projects (Grant Number: RM124190D66C576E).
\balance
\bibliographystyle{latex8}
\bibliography{bibliography}

\begin{thebibliography}{10}\setlength{\itemsep}{-1ex}\small

\bibitem{doi:10.1177/10692509251339913}
D.~Avola, F.~Bruni, G.~L. Foresti, D.~Pannone, and A.~Ranaldi.
\newblock Digital shielding for cross-domain wi-fi signal adaptation using
  relativistic average generative adversarial network.
\newblock {\em Integrated Computer-Aided Engineering}, pages 1--17, 2025.

\bibitem{doi:10.1142/S0129065722500150}
D.~Avola, M.~Cascio, L.~Cinque, A.~Fagioli, and G.~L. Foresti.
\newblock Human silhouette and skeleton video synthesis through wi-fi signals.
\newblock {\em International Journal of Neural Systems}, 32(05):2250015, 2022.

\bibitem{9730862}
D.~Avola, M.~Cascio, L.~Cinque, A.~Fagioli, and C.~Petrioli.
\newblock Person re-identification through wi-fi extracted radio biometric
  signatures.
\newblock {\em IEEE Transactions on Information Forensics and Security},
  17:1145--1158, 2022.

\bibitem{AVOLA2024103643}
D.~Avola, L.~Cinque, M.~{De Marsico}, A.~Fagioli, G.~L. Foresti, M.~Mancini,
  and A.~Mecca.
\newblock Signal enhancement and efficient dtw-based comparison for wearable
  gait recognition.
\newblock {\em Computers \& Security}, 137:103643, 2024.

\bibitem{BOWYER2008281}
K.~W. Bowyer, K.~Hollingsworth, and P.~J. Flynn.
\newblock Image understanding for iris biometrics: A survey.
\newblock {\em Computer Vision and Image Understanding}, 110(2):281--307, 2008.

\bibitem{Butterworth1930}
S.~Butterworth.
\newblock On the {{Theory}} of {{Filter Amplifiers}}.
\newblock {\em Experimental Wireless \& the Wireless Engineer}, 7:536--541,
  1930.

\bibitem{hampel_ID}
L.~Davies and U.~Gather.
\newblock The identification of multiple outliers.
\newblock {\em Journal of the American Statistical Association},
  88(423):782--792, 1993.

\bibitem{RSSI_reliability_indoorloc}
Q.~Dong and W.~Dargie.
\newblock Evaluation of the reliability of rssi for indoor localization.
\newblock In {\em International Conference on Wireless Communications in
  Underground and Confined Areas (ICWCUCA)}, pages 1--6, 2012.

\bibitem{ESP32_tool}
S.~M. Hernandez and E.~Bulut.
\newblock {Lightweight and Standalone {IoT} Based {WiFi} Sensing for Active
  Repositioning and Mobility}.
\newblock In {\em International Symposium on {"}A World of Wireless, Mobile and
  Multimedia Networks{"} (WoWMoM) (WoWMoM)}, pages 277--286, 2020.

\bibitem{EdgeWiFiSensing2022}
S.~M. Hernandez and E.~Bulut.
\newblock Wifi sensing on the edge: Signal processing techniques and challenges
  for real-world systems.
\newblock {\em IEEE Communications Surveys \& Tutorials}, 25(1):46--76, 2023.

\bibitem{10.1145/3310194}
Y.~Ma, G.~Zhou, and S.~Wang.
\newblock Wifi sensing with channel state information: A survey.
\newblock {\em ACM Computing Surveys}, 52(3):1--35, 2019.

\bibitem{Gait_hampel_pca}
R.~Ou, Y.~Chen, and Y.~Deng.
\newblock Wiwalk: Gait-based dual-user identification using wifi device.
\newblock {\em IEEE Internet of Things Journal}, 10(6):5321--5334, 2023.

\bibitem{7380482}
V.~J. Rathod, N.~C. Iyer, and M.~S. M.
\newblock A survey on fingerprint biometric recognition system.
\newblock In {\em International Conference on Green Computing and Internet of
  Things (ICGCIoT)}, pages 323--326, 2015.

\bibitem{RSSI_indoorloc}
S.~Sadowski and P.~Spachos.
\newblock Rssi-based indoor localization with the internet of things.
\newblock {\em IEEE Access}, 6:30149--30161, 2018.

\bibitem{Phase_cal1}
S.~Sen, B.~Radunovic, R.~R. Choudhury, and T.~Minka.
\newblock You are facing the mona lisa: Spot localization using phy layer
  information.
\newblock In {\em ACM International Conference on Mobile Systems, Applications,
  and Services (MobiSys)}, page 183–196, 2012.

\bibitem{8528404}
J.~P. SINGH, S.~JAIN, S.~ARORA, and U.~P. SINGH.
\newblock Vision-based gait recognition: A survey.
\newblock {\em IEEE Access}, 6:70497--70527, 2018.

\bibitem{vaswani2023attentionneed}
A.~Vaswani, N.~Shazeer, N.~Parmar, J.~Uszkoreit, L.~Jones, A.~N. Gomez, L.~u.
  Kaiser, and I.~Polosukhin.
\newblock Attention is all you need.
\newblock In {\em Advances in Neural Information Processing Systems (NIPS)},
  pages 1--11, 2017.

\bibitem{WiPIN}
F.~Wang, J.~Han, F.~Lin, and K.~Ren.
\newblock Wipin: Operation-free passive person identification using wi-fi
  signals.
\newblock In {\em IEEE Global Communications Conference (GLOBECOM)}, pages
  1--6, 2019.

\bibitem{Phase_cal3}
X.~Wang, L.~Gao, and S.~Mao.
\newblock Phasefi: Phase fingerprinting for indoor localization with a deep
  learning approach.
\newblock In {\em IEEE Global Communications Conference (GLOBECOM)}, pages
  1--6, 2015.

\bibitem{CSI_RSSI}
Z.~Yang, Z.~Zhou, and Y.~Liu.
\newblock From rssi to csi: Indoor localization via channel response.
\newblock {\em ACM Computing Surveys}, 46(2):1--32, 2013.

\bibitem{WiWho_gait_decisionTree}
Y.~Zeng, P.~H. Pathak, and P.~Mohapatra.
\newblock Wiwho: Wifi-based person identification in smart spaces.
\newblock In {\em ACM/IEEE International Conference on Information Processing
  in Sensor Networks (IPSN)}, pages 1--12, 2016.

\bibitem{7536315}
J.~Zhang, B.~Wei, W.~Hu, and S.~S. Kanhere.
\newblock Wifi-id: Human identification using wifi signal.
\newblock In {\em International Conference on Distributed Computing in Sensor
  Systems (DCOSS)}, pages 75--82, 2016.

\bibitem{9516988}
Y.~Zhang, Y.~Zheng, K.~Qian, G.~Zhang, Y.~Liu, C.~Wu, and Z.~Yang.
\newblock Widar3.0: Zero-effort cross-domain gesture recognition with wi-fi.
\newblock {\em IEEE Transactions on Pattern Analysis and Machine Intelligence},
  44(11):8671--8688, 2022.

\bibitem{10.1145/954339.954342}
W.~Zhao, R.~Chellappa, P.~J. Phillips, and A.~Rosenfeld.
\newblock Face recognition: A literature survey.
\newblock {\em ACM Computing Surveys}, 35(4):399–458, 2003.

\bibitem{phase_noise}
A.~Zhuravchak, O.~Kapshii, and E.~Pournaras.
\newblock Human activity recognition based on wi-fi csi data -a deep neural
  network approach.
\newblock {\em Procedia Computer Science}, 198:59--66, 2022.

\end{thebibliography}
\end{document}